\documentclass[10pt,twocolumn,letterpaper]{article}

\usepackage[pagenumbers]{wacv} 

\usepackage{graphicx}
\usepackage{amsmath}
\usepackage{amssymb}
\usepackage{booktabs}
\usepackage[normalem]{ulem}
\usepackage[accsupp]{axessibility}  

\usepackage[pagebackref,breaklinks,colorlinks]{hyperref}

\usepackage[capitalize]{cleveref}
\crefname{section}{Sec.}{Secs.}
\Crefname{section}{Section}{Sections}
\Crefname{table}{Table}{Tables}
\crefname{table}{Tab.}{Tabs.}

\begin{document}

\title{Enhancing Multimodal Compositional Reasoning of Visual Language Models with Generative Negative Mining}

\author{{Ugur Sahin$^*$}$^{1}$
\qquad
{Hang Li$^*$}$^{2,3}$
\qquad
{Qadeer Khan}$^{1,4}$
\qquad
{Daniel Cremers}$^{1,4}$
\qquad
{Volker Tresp}$^{2,4}$\\
{$^{1}$Technical University of Munich\qquad $^{2}$LMU Munich\qquad $^{3}$Siemens AG}\\ {$^{4}$Munich Center for Machine Learning}\\
}

\maketitle
\def\thefootnote{*}\footnotetext{Equal contribution.}\def\thefootnote{\arabic{footnote}}

\begin{abstract}

Contemporary large-scale visual language models (VLMs) exhibit strong representation capacities, making them ubiquitous for enhancing image and text understanding tasks. They are often trained in a contrastive manner on a large and diverse corpus of images and corresponding text captions scraped from the internet. Despite this, VLMs often struggle with compositional reasoning tasks which require a fine-grained understanding of the complex interactions of objects and their attributes. This failure can be attributed to two main factors: 1) Contrastive approaches have traditionally focused on mining negative examples from existing datasets. However, the mined negative examples might not be difficult for the model to discriminate from the positive. An alternative to mining would be negative sample generation 2) But existing generative approaches primarily focus on generating hard negative texts associated with a given image. Mining in the other direction, i.e., generating negative image samples associated with a given text has been ignored.  To overcome both these limitations, we propose a framework that not only mines in both directions but also generates challenging negative samples in both modalities, i.e., images and texts. Leveraging these generative hard negative samples, we significantly enhance VLMs' performance in tasks involving multimodal compositional reasoning. Our code and dataset are released at \url{https://ugorsahin.github.io/enhancing-multimodal-compositional-reasoning-of-vlm.html}.

\end{abstract}

\begin{figure}
    \centering
    \includegraphics[width=\linewidth]{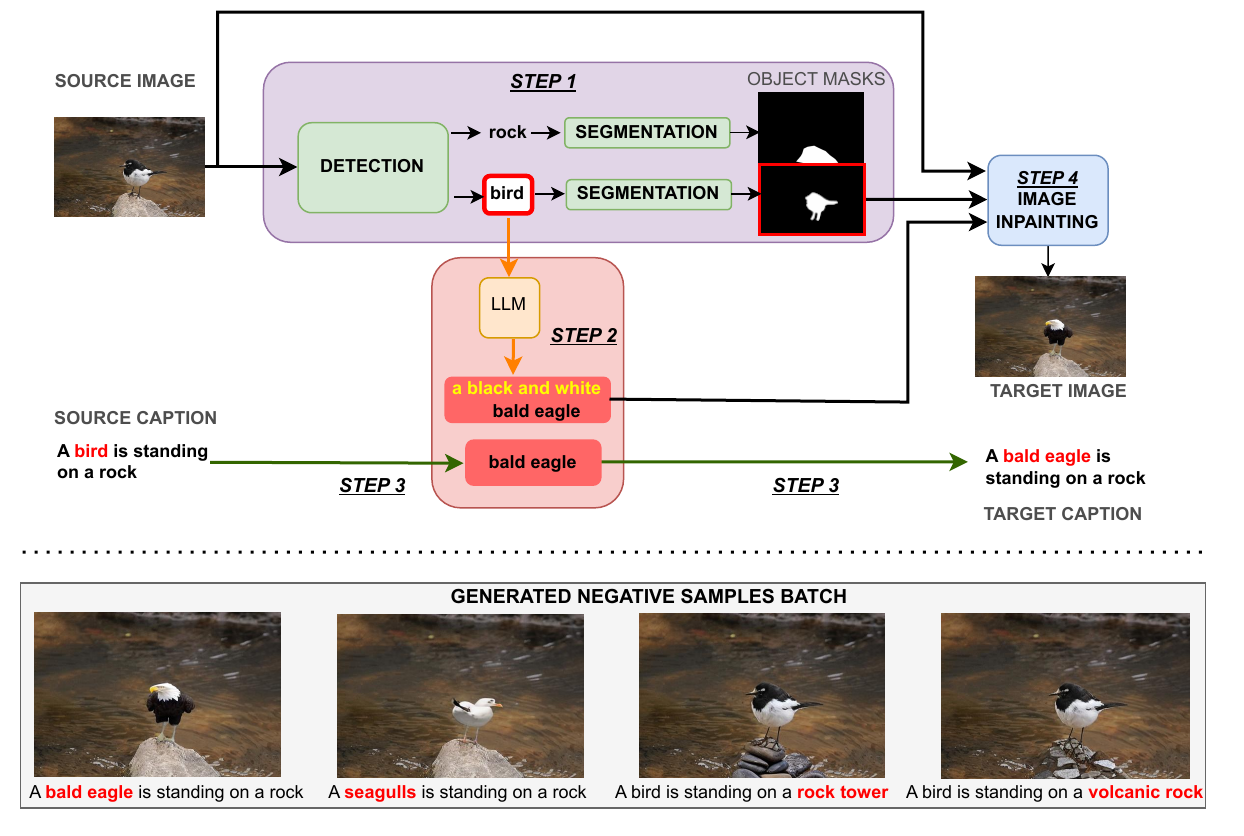}
    \caption{\textbf{Top:} Gives the overview of our proposed generative approach for image-text synthesis from a given source image and a corresponding caption. \textit{Step 1:} The source image is first passed through a detection and segmentation algorithm to identify all the relevant objects in the scene (bird and rock) and also create independent masks of these objects (See Subsection \ref{subsubsec:step1}). The remaining steps in this figure focus on the bird object. \textit{Step 2:} A large language model (LLM) then takes the detected objects to create 1) an alternate representation of that object (bald eagle) 2) A more fine-grained and descriptive representation of the same object (a black and white bald eagle) (See Subsection \ref{subsubsec:step2}). \textit{Step 3:}. The source caption is replaced with an alternate representation to produce the target caption. \textit{Step 4:} The original mask of the object and the descriptive alternate caption are fed to an inpainting algorithm to replace the \textit{bird} with \textit{a black and white bald eagle} in the source image to produce the target image (See Subsection \ref{subsubsec:step4}). \textbf{Bottom:} Shows a batch of some other generated variations of the same source image.}
    \label{fig:introduction}
\end{figure}

\section{Introduction}
\label{sec:intro}

Contrastive learning has been demonstrated to be an effective and popular technique for training large large-scale visual language models\cite{radford2021learning, li2021align,zhang2021vinvl}. This is due to the availability of a large corpus of images and text reaching an order of millions of samples that can readily be scraped from the internet\cite{schuhmann2021laion, schuhmann2022laion}. Training on human-curated supervised data of a similar scale would be infeasible due to the sheer amount of effort required for annotation. Meanwhile, training on the limited supervised data would not yield results of performance comparable to that of contrastive methods trained on data of much higher magnitude.  In fact, this contrastive pretraining on a large corpus of data has led to enhanced image and text representations that benefit downstream tasks including image and text retrieval\cite{yuksekgonul2023when,momeni2023verbs}, text generation\cite{mokady2021clipcap,li2022blip,li2023blip}, and image generation\cite{rombach2022high,ramesh2022hierarchical}.

Despite the impressive results VLMs have achieved on the above tasks, one challenging problem that still remains is their limited compositional ability\cite{thrush2022winoground,yuksekgonul2023when,ma2023crepe}. Compositionality refers to the challenge where the samples have significantly different semantic scene depictions despite similar textual representations. For e.g. the two sentences  \textit{1) a black dog with a white cat} and  \textit{2) a white dog with a black cat} may appear to be textually similar but have very different scene depictions. While humans can easily discern the context between the two sentences, VLMs tend to struggle, posing a significant challenge in regards to this compositional reasoning~\cite{schiappa2023probing,thrush2022winoground}. This is further exacerbated when words in the sentences are exactly the same but differ only in order, as is the case in the example described above. \cite{thrush2022winoground} proposed a new dataset to specifically evaluate this compositional reasoning of various VLMs. They showed that these VLMs tend to struggle with compositionality. 
A plausible explanation is given in recent work which identified that VLMs are prone to exploiting shortcut strategies~\cite{geirhos2020shortcut}. Given a caption, the VLM may choose to focus on only a certain region among the rich scene representation while ignoring other objects in the scene. For e.g. given the source caption in Fig. \ref{fig:introduction}: \textit{a bird is standing on a rock} may decide to only focus on the bird and the rock while completely ignoring the background and placing less emphasis on the bird species or type of rock.  This tends to happen because in the usual contrastive learning setting, the negative samples are already significantly different from one another. Hence, VLMs only need to detect these major differences, instead of truly understanding the complex structure of the entire scene~\cite{chen2020cops}. 

To train a model to truly compositionally reason about the scene and text, we would like to mine for hard negative samples. This hard negative mining is among the promising directions to tackle this problem~\cite{yuksekgonul2023when,li2021align,robinson2020contrastive,huang2023structure,momeni2023verbs}. It includes finding examples with minimal changes in the text or image but yielding different contexts. The bottom part of Fig. \ref{fig:introduction} shows four negative samples of the source data point. Note that the caption and image of the negative samples have a subtle difference from the source but it completely changes the context (bird species, rock formation). Such negative samples capture a more fine-grained representation of the image and text content. The model is now forced to learn the subtle differences between for e.g. a seagull and a bald eagle or volcanic rock and a tower of rocks. Just being aware that some bird is sitting on some rock would not be enough for the model. It has to additionally focus on the bird species and type of rock formation. However, such hard negative samples with subtle differences in text and images may not necessarily exist. Therefore, how do we mine for such non-existent hard negative samples in both modalities, i.e., text and images? For text, most existing works on negative mining augment the textual descriptions~\cite{doveh2023teaching,yuksekgonul2023when,huang2023structure}. But how can we ensure that the generated sentences are even linguistically meaningful? Moreover, how do we mine hard negative samples in the image space?

Motivated by recent advances in image understanding and generation~\cite{ho2020denoising, rombach2022high, ramesh2022hierarchical, li2023dall, li2022blip}, we propose a framework to generate negative images to facilitate contrastive learning. Specifically, recent development in image understanding models such as SAM enables a reliable segmentation of objects from a complex scene. Moreover, large image generation models such as Stable Diffusion (SD) can convert text descriptions into images. The inpainting mode of SD allows it to modify part of the image while keeping the remaining part unchanged. As shown in the upper part of Fig. \ref{fig:introduction} with blue arrows, we are able to edit the original images with minimal changes in the pixel space thus constituting hard-to-discriminate image examples. This is done by replacing the word \textit{bird} with the prompt \textit{bald eagle}. Generating images similar in category (e.g., bird) but different in appearance for a large number of samples would be a tedious process for a human. Therefore, we automate the process using LLMs to generate such prompts. These LLMs automatically propose alternative concepts to replace the original words in the caption without losing the linguistic meaning. 
As shown in Fig. \ref{fig:introduction}, the LLM changes the word \textit{bird} into a specific category of \textit{bald eagle}. In the end, we obtain a batch of hard negative examples at the bottom of Fig. \ref{fig:introduction}. To match the left-most image from four texts with minimal word changes, the model needs to encode fine-grained bird and rock information from the image and texts.

We conducted extensive experiments using the dataset generated by our proposed method to demonstrate the power of our framework. In this regard, the contributions of our framework are summarized as follows:
\begin{itemize}
    \item We show that our method which uses negative sample generation improves VLM performance on a wide range of benchmarks meant to assess compositional visual-language reasoning.  These include Winoground, ARO, CREPE, and VL-Checklist.
    
    \item We release our dataset which is comprised of fine-grained object differences and attribute changes in the images and text.  Such subtle differences in the dataset make it challenging for pre-trained state-of-the-art visual language models to correctly compositionally reason about the data points. The supplementary material contains a subset of our dataset. The complete dataset is accessible at our project page \url{https://ugorsahin.github.io/enhancing-multimodal-compositional-reasoning-of-vlm.html}.

\end{itemize}

\section{Related Work}

\vspace{0.1cm}
\noindent\textbf{Contrastive pre-training of VLMs}
Contrastive Pre-training of large-scale models trained together on both vision and language modalities have shown superior representation~\cite{li2021align,radford2021learning} and zero-shot transfer ability~\cite{huang2023structure}, leading to success on a wide spectrum of related tasks~\cite{li2022blip,mokady2021clipcap,rombach2022high}. Due to the significant amount of image-text data crawled from the internet~\cite{schuhmann2021laion,schuhmann2022laion}, the unsupervised contrastive learning paradigm stands out as a primary approach to pre-train VLMs~\cite{radford2021learning,jia2021scaling,yao2021filip}. Contrastive learning relies on negative examples which train models to discriminate between them and the positive examples. If the negative examples are significantly different from the positive, the model can easily discriminate between the two. In contrast, if the negative and positive samples are similar, the model can learn to correctly discriminate only if it understands the subtle, fine-grained differences between the two. Such hard negative samples provide the model with greater predictive power. Inspired by metric learning~\cite{robinson2020contrastive}, hard negative mining based on learned embeddings became a popular approach to improve contrastive learning~\cite{robinson2020contrastive}, with different methods for mining negative samples being proposed~\cite{huynh2022boosting,zhang2021understanding, wang2023boosting,chuang2020debiased,radenovic2023filtering, chen2020counterfactual}.  To address the limitation that certain negative examples are hard to find in existing datasets, recent works have rather explored synthesizing negative text, and hard negative captions on standard image-text datasets~\cite{lin2014microsoft} by word shuffling~\cite{yuksekgonul2023when}, negative verbs~\cite{momeni2023verbs}, negative text augmentation~\cite{doveh2023dense,doveh2023teaching, schiappa2023probing, huang2023structure, momeni2023verbs}. Compared to these approaches, our method focuses on mining hard negatives from both image and text domains, leveraging large-scale generative language and vision models.

\vspace{0.1cm}
\noindent\textbf{Benchmarking Visual-linguistic Compositional reasoning}
Compositional reasoning is the ability to understand complex scenes and text with diverse structures~\cite{thrush2022winoground}. This includes for e.g. the capacity to discern between sentences with the same words but in a different order, or a scene with the same objects but slightly different colors, etc. There are many datasets~\cite{zhao2022vl, thrush2022winoground, diwan2022winoground, ma2023crepe} used for benchmarking different aspects of compositional reasoning. For e.g. Winoground~\cite{thrush2022winoground} tests for rich structures in text order, CREPE~\cite{ma2023crepe} for constituting objects, their relations, and attributes, ARO~\cite{yuksekgonul2023when} for shuffled word order. These benchmarks demonstrated that most SOTA VLMs showed poor performance when probed for compositional reasoning. Our method on the other hand is capable of understanding the subtle visual-lingustic cues thereby demonstrating superior performance.

\vspace{0.1cm}
\noindent\textbf{Additional Image and Text Data Generation}
Synthetic image generation using text-to-image models has proven effective in various computer vision tasks such as image classification~\cite{he2022synthetic}, object detection~\cite{wu2023diffumask,ni2022imaginarynet}, image captioning~\cite{trabucco2023effective}, and contrastive learning~\cite{cascante2023going}. Generated images can complement existing datasets with a diverse set of images that may not be present in the existing datasets, enriching the overall range of available visual examples. \cite{cascante2023going} propose to improve contrastive learning with synthetic image generation, which is probably the closest to our work. However, it generates synthetic images from scratch, whereas we edit realistic images from a human-curated dataset. Generating additional text samples from LLMs is a very promising direction. LLMs such as ChatGPT and LLaMA exhibit well modeling of language structure\cite{brown2020language, garg2022can} and thus can be utilized to manipulate text to enrich the text samples~\cite{doveh2023teaching}. For example, \cite{fan2023improving} proposes to rewrite texts in COCO to improve contrastive image-text pertaining~\cite{kumar2020data}. Similarly, we utilize LLM to generate contrastive text samples for detected objects in the image. Then we utilize text-to-image models to edit the original image to obtain negative examples.

\section{Method}

This section first outlines our data generation pipeline that leverages the latest state-of-the-art LLMs and multimodal generative models for high-quality sample generation (See Fig. \ref{fig:introduction}). We then present the finetuning framework that exploits our new dataset of hard negative examples to enhance the compositional reasoning abilities of VLMs.

\begin{figure*}[htp]
    \centering
    \includegraphics[width=\linewidth]{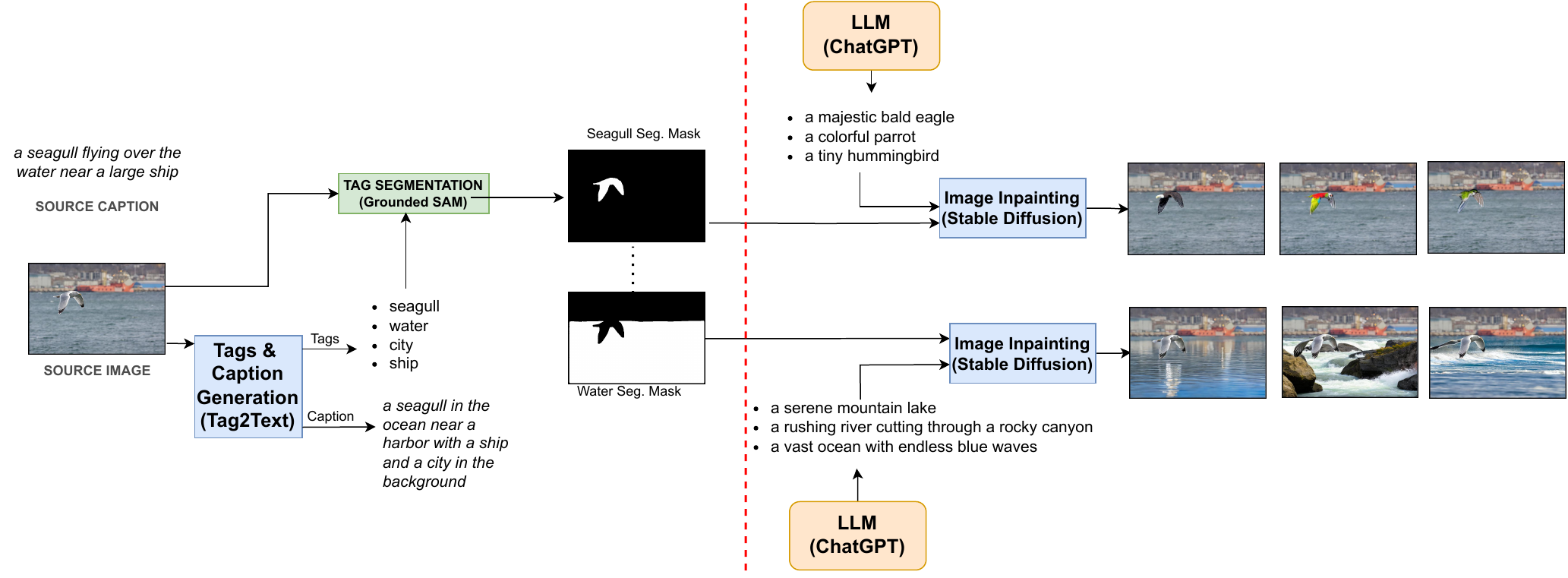}
    \caption{Overview of data generation pipeline. \textbf{Left:} The portion to the left of the red dotted line demonstrates the process for determining segmentation masks of all objects in the scene, which is elaborated in Subsection \ref{subsubsec:step1}. The Tag2Text model is first utilized to generate a list of tags for all objects in the scene. Segmentation masks from the source image are then created for all the individual tags (Masks for the seagull and water tags are shown). Note that the human-annotated source caption may not contain all the identified tags, e.g., city. Therefore, Tag2Text also generates a caption for the source image to encompass all the detected objects. {The replacement of concepts for a new caption generation is explained in Subsection \ref{subsubsec:step3}.} \textbf{Right:} The portion to the right of the red dotted line figure corresponds to the process of generating images having subtle variations from the source image, as explained in Subsection \ref{subsubsec:step4}. For this, we use the Stable Diffusion model which takes the segmentation masks along with the fine-grained description of objects with which the masks are replaced. The new descriptions are produced using ChatGPT.}
    \label{fig:method_image}
\end{figure*}

\subsection{Hard Negative Example Generation}

Our framework can be used to enhance the richness of any image-caption pair dataset. For our experiments, we use data generated based on the human-curated image-text pair COCO dataset, {but our approach can be extend to other datasets.} In the following, we describe the core components of our generation pipeline. The main objective is to generating challenging negative examples which modifies local semantics of the scene and preserves the main context.

\subsubsection{Image Analysis and Object Extraction}\label{subsubsec:step1}
To accurately identify the regions of an image that need modification, we utilize a comprehensive annotation approach to decompose the scene into its constituent parts. Firstly we utilize off-the-shelf image-to-text models, specifically Tag2Text~\cite{huang2023tag2text}, for object detection and caption generation. As shown on the left side of Fig. \ref{fig:method_image}, the Tag2Text model outputs a list of detected object labels in the image, along with a descriptive caption summarizing the entire scene. The descriptive caption is needed to ensure that all the identified objects have been covered in the caption. 

However, Tag2Text lacks precise object localization capabilities and cannot demarcate precise object boundaries in the scene. To circumvent this issue, we integrate a segmentation model, such as Grounded-SAM~\cite{kirillov2023segany,liu2023grounding} into our framework. The segmentation model takes as input both the image and a label of one of the objects in the same image. Its output generates a binary mask that highlights the corresponding object region within the image.

\subsubsection{Concept Augmentation Using LLM}\label{subsubsec:step2}
\begin{figure}[htp]
    \centering
    \includegraphics[width=\linewidth]{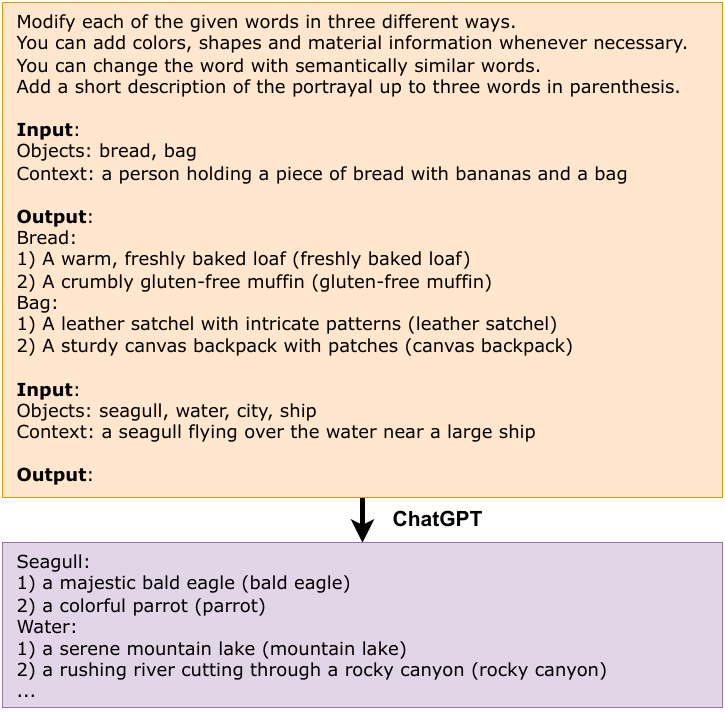}
    \caption{Text variation generation by LLM, explained in Subsection \ref{subsubsec:step2}. Following the pattern defined in the prompt which changes the objects bird and rock into different instances with rich attributes, LLM completes the text for the test sample, i.e., transforming water into different types of water.}
    \label{fig:method_llm}
\end{figure}

A detected object is transformed into a similar concept. Our aim focuses on modifying the object's appearance, attributes, and categories while keeping other things and the overall context the same. This includes transforming an object into a more fine-grained instance with richer attributes (e.g., transforming a house to a Victorian one with a wooden entrance), or modifying the background into different environments (transforming the sky into rocky mountains). For that, we resort to LLMs, such as open-sourced LLaMA~\cite{touvron2023llama} and ChatGPT which offer impressive possibilities due to their in-context learning capacity~\cite{touvron2023llama}. Given a few examples and a test sample, LLMs generate the output that adheres to the structures implied by the set of given examples. For instance, as illustrated in Fig. \ref{fig:method_llm}, the input prompt presents an example where the word \textit{bread} is modified into \textit{freshly baked loaf}.
When the LLM is prompted with a test case \textit{water}, it generates a \textit{mountain lake} that follows a similar modification pattern. To enhance the generation, a source caption is fed to the LLM as context information, encouraging the generation of object variations that are more compatible with the background. Additionally, the prompt is manually designed to guide the LLM toward producing the desired output, i.e., instructions such as \textit{using a maximum of three words} give more control over the style of the generated outputs. Further, the LLM is instructed to generate keywords summarizing the detailed descriptions. The detailed descriptions are used in image editing, whereas the keywords are used to replace the caption. This strategy ensures a relatively precise image caption, as well as adding more fine-grained details to the visual scene in the image generation stage. Our approach can automate the entire process by conveniently utilizing the ChatGPT API.

Note that this generation process is open-ended and can synthesize an arbitrary number of data samples. Being able to train on a large dataset is where the power of contrastive learning comes from. This is as opposed to supervised methods whose training is restricted to the number of labeled samples which are expensive and tedious to collect.

\subsubsection{Caption Editing} \label{subsubsec:step3}
{
We replace the object in the original source caption with the newly generated phrase. For example, in Fig. \ref{fig:method_image}, we replace the seagull with a bald eagle to create a caption for the newly generated image. Specifically, the source human-annotated caption \textit{a seagull flying over the water near a large ship} is changed into \textit{a bald eagle flying over the water near a large ship}. However, Tag2Text may produce tags, e.g., \textit{city}, that are not presented in the source caption. For that, the caption produced by the Tag2Text model is used to generate the ground truth caption for the modified image. For example, if the tag \textit{city} is augmented to \textit{historic town} by the LLM (explained in Subsection \ref{subsubsec:step2}), we label the augmented image with caption: \textit{a seagull in the ocean near a harbor with a ship and a historic town in the background}.} We label the generated image with our edited caption as a ground-truth image-text pair.

\subsubsection{Image Editing} \label{subsubsec:step4}
To enable fine-grained modification of an image region, we adopt the concept of image inpainting for transforming the original object in the image into the target object. In this scenario, image inpainting involves removing a specific region and filling it with content that seamlessly integrates with the image's context while considering the input information. The inpainting model takes multiple inputs, as illustrated on the right side in Fig. \ref{fig:method_image}. One of the inputs is the binary mask that identifies the object region that is desired to be replaced with a new object in the source image (See Subsection \ref{subsubsec:step1}). The other important input to the inpainting model is the object description obtained from the LLM, which indicates the target object we want the region to be changed into (See Subsection \ref{subsubsec:step2}). The model's output is a modified image that aims to replace the content with the desired object description while the remaining parts are unchanged. This process ensures that the modified images are realistic and similar to the original images.

\subsubsection{Filtering}
For each image, we randomly sample $M$ objects to input to the LLM, which subsequently generates $K$ text variations for each selected object. We employ filters to eliminate certain examples, e.g., wrong segmentation mask, missing parts in segmentation, confusion due to multiple objects, or the text is not descriptive enough. To address these issues, we utilize BLIP's~\cite{li2022blip} ITM head, which outputs a confidence value if a given image and text pair matches. Notice that generated image might be too noisy that it barely changes even in pixels, to filter the generated images, we first calculate the standard deviation across them followed by averaging. Then we calculate the average of standard deviance within channels. Images are removed if the difference is smaller than a threshold value. {More implementation details are in Appendix A.}

\subsection{Finetuning Framework}

\subsubsection{Preliminary: Contrastive Loss}
The CLIP model operates on a pair of image $I$ and text $T$, encoding them separately into embedding space $\mathbb{R}^d$, denoted as $e_I=\mathcal{E}_I(I)$ and $e_T=\mathcal{E}_T(T)$. The image-text similarity score is computed as 
$$S(T,I)=\exp \bigl( \frac{ e_T^T e_I/\tau }{|| e_T ||^2 || e_I ||^2} \bigl),$$ where temperature $\tau$ is a learnable parameter. During the training process, we sample a batch of $N$ pairs of images and texts from the training dataset. The training objective aims to maximize the similarity between matched pairs and minimize the similarity between unmatched pairs. This is achieved through the contrastive loss, formulated as $$\mathcal{L} = \sum_i^N log\bigl( \frac{S(T_i,I_i)}{\sum_j^N S(T_i, I_j)} \bigl) + log\bigl(  \frac{S(T_i,I_i)}{\sum_k^N S(T_k, I_i)} \bigl).$$
The first part of the loss ensures that for each text $T_i$, we increase the similarity to its paired image $I_i$ while decreasing the similarity to the remaining images in the batch. Similarly, the second part iterates over texts for a sampled image, encouraging similarity to the paired text and discouraging similarity to other texts in the batch. By employing this loss function, we finetune CLIP on our dataset.

\subsubsection{Mixing of Hard Negative Examples}
As our data generation method is unsupervised, it may produce images that do not correspond well with the expected text. To mitigate the negative impact of noise in the dataset, we combine the generated samples with the original human-annotated COCO image-text pairs. Moreover, these image-text pairs, despite being of a smaller scale compared to the dataset used for pertaining CLIP, serve as a valuable resource to prevent the model from overfitting. We employ a simple strategy to sample a batch. For a batch, we sample $rN$ pairs from our generated data and $(1-r)N$ pairs from the original COCO dataset. These pairs are then concatenated to form a single batch of size $N$. 

\section{Experiments}
This section describes the experimental setups, including the datasets used for evaluation, implementation details of the finetuning pipeline, evaluation metrics, and baseline models for a comprehensive comparison.

\vspace{0.1cm}
\noindent\textbf{Datasets}
We evaluate our model on composition-oriented benchmarks of different scales and different compositional aspects. The following benchmarks are included in our experiments.  1) Winoground is a hand-crafted dataset of 800 image-text pairs, for each set of two texts, the texts have exactly the same words but with different word orders, the texts are mapped to two visually distinct images. 2) ARO has more than 50,000 test images paired with automatically built text examples with changed attributes, relationships, and word order, leveraging VG~\cite{krishna2017visual}, COCO~\cite{lin2014microsoft}, and Flickr~\cite{young2014image}. 3) CREPE introduces new negative texts for existing images in CC-12M~\cite{changpinyo2021conceptual}, YFCC-5M~\cite{thomee2016yfcc100m}, LAION-400M~\cite{schuhmann2021laion}, where the number of changed words in the text is gradually increased, treated as different levels of complexity. For ARO and CREPE, their texts are generated with unique methods which are not covered in our training data. This makes them perfect candidates to verify the generalization ability of our approach. For the ablation study, we primarily perform experiments on Winoground, as it is manually verified and more challenging~\cite{diwan2022winoground}, since each text in this dataset has a corresponding hard negative image with complex semantics.

Our training dataset is generated based on the COCO dataset, which has a training split with 110k image-text pairs. In our experiments, we created variations for 12.656 unique images, where for each image, we selected approximately three objects on average and generated four text variations for each object. After filtering out low-quality generations, we ended up with 82.010 image and text pairs. We generate our test set from the COCO Karpathy test split~\cite{karpathy2015deep} with 5k image-text pairs. Additionally, we manually verify the generation to better inform the model selection and more importantly, serve as sources for the community to facilitate visual language research. For our test set, we have 278 unique images with different image-text pairs for each of them due to our manual filtering. In the end, we obtained 122 images with 4 variations for each image, 139 images with 3 variations, and 17 images with 2 variations.

\vspace{0.1cm}
\noindent\textbf{Implentation Details}
For dataset generation, we utilize the public implementation of Tag2Text\footnote{https://tag2text.github.io/}, Grounded-SAM\footnote{https://github.com/IDEA-Research/Grounded-Segment-Anything}, and Stable Diffusion\footnote{https://github.com/CompVis/stable-diffusion}. For finetuning, we follow the strategies in similar work~\cite{yuksekgonul2023when, momeni2023verbs} and combine our generated sample with human-annotated labels. We set the ratio of real and synthetic data as $r=0.5$. We utilize the OpenAI CLIP ViT/B-32 architecture. We use a batch size of 400, a learning rate of 1e-6, and a weight decay of 0.2, and fine-tune the model for 20 epochs. We employ the default image augmentation techniques that were used during pretraining CLIP. The experiments are conducted on an Nvidia A10G GPU with 24GB memory.

\vspace{0.1cm}
\noindent\textbf{Evaluation}
We adopt the evaluation metric for different datasets, which are basically formulated as image-to-text and text-to-image retrieval tasks. For Winoground, we report image score, text score, and group score, meaning that the model should correctly choose the text among the two text candidates for each image. For CREPE, we report the hits@1 image-to-text retrieval score on the productivity set with complexity ranging from 4 to 12. For ARO, we employ their evaluation metric and report the mean of the performance for each subcategory.

\section{Results}
In this section, we present the comprehensive experimental results of our proposed method in comparison to the baseline across a diverse range of tasks. Moreover, we further verify the effectiveness of our generated dataset through an extensive ablation study.

\begin{table}[htpb]
    \centering
    \resizebox{0.85\linewidth}{!}{
    \begin{tabular}{c|ccc}\toprule
        Model & Text Score & Image Score & Group Score  \\ \hline
        CLIP & 30.75 & 11.0 &  8.75 \\
        Ours & \textbf{34.25} & \textbf{12.5} & \textbf{10.0} \\ \hline
        Relative Gains & +11.1\% & +13.6\%& +14.2\% \\
       \toprule
    \end{tabular}}
    \caption{Comparison of our method with CLIP on Winoground benchmarks. We report the text score, image score, and group score which measure if the model can correctly match a text for an input image, or vice versa. The best performance is shown in \textbf{bold}. Our finetuned CLIP surpasses the baseline model by a substantial margin.}
    \label{tab:winoground}
\end{table}

\begin{table}[htp]
    \centering
    \scriptsize
    \begin{tabular}{c|ccc|ccc}\toprule
        & \multicolumn{3}{c|}{Compositional (171)} & \multicolumn{3}{c}{Complex (78)} \\\hline
       CLIP & 31.58&            11.70&           9.36& 23.08&            6.41&           3.85\\
        Ours & \textbf{38.01}&           \textbf{14.62}&           \textbf{10.53} & \textbf{29.49}&  \textbf{8.97}&   \textbf{6.41}\\ \hline
        Gains &  +22.5\% & +27.2\% & +12.5\% & +23.9\% & +39.9\% & +66.5\% \\ \toprule
        & \multicolumn{3}{c|}{Unusual Image (56)} & \multicolumn{3}{c}{Unusual Text (50)} \\\hline
        CLIP & 26.79&           \textbf{8.93}& 5.36& \textbf{34.0}&            \textbf{14.0}&            \textbf{10.0}\\
        Ours & \textbf{28.57}&           \textbf{8.93}&           \textbf{8.93}& 30.0&            10.0&            \textbf{10.0}\\ \hline
        Gains & +6.7\% & 0.0\% & +66.3\% & -11.8\% & -28.5\% & 0.0\% \\ \toprule
    \end{tabular}
    \caption{Comprehensive analysis of method performance on Winoground subsets~\cite{diwan2022winoground} which evalute distinct reasoning abilities. Numbers in the parenthesis indicate the number of samples in that subcategory. Our model excels in compositional reasoning tasks, while it may face challenges in tasks that require an understanding of unusual text which entails background knowledge.}
    \label{tab:winoground-sub}
\end{table}

\subsection{Evaluation on Visuo-Linguistic Benchmarks}
Table \ref{tab:winoground}, \ref{tab:winoground-sub}, \ref{tab:aro} provide a detailed comparison of our finetuned CLIP model on our generated hard negatives with the released COCO checkpoints. We evaluate the model performance across a wide range of visual language reasoning tasks with various aspects of reasoning ability. Our findings demonstrate that our method outperforms CLIP by a substantial margin in the majority of these tasks. Specifically, in Tab. \ref{tab:winoground}, we highlight the significant improvement achieved by our method in terms of complex image text matching tasks. Furthermore, we achieved a relative improvement of 22.5\% in text score and 27.2\% in image score on the compositional split filtered by~\cite{diwan2022winoground}. This subset ensures image-text matching with only compositional ability, instead of other abilities such as visual difficulty. We report the performance of our method on detailed subcategories of Winoground with most test cases in Tab. \ref{tab:winoground-sub}.{Our approach exhibits lower performance on the unusual text subset, which emphasizes understanding the nuanced meaning of the text. For example, matching \textit{the brave in the face of fear} with an image that depicts a small cub confronting a fierce lion, is challenging for our approach. The presence of repetitive text samples in our augmented dataset may impact the finetuning of the text encoder (see Appendix B)}.

\begin{table}[htp]
    \centering
    \resizebox{\linewidth}{!}{
    \begin{tabular}{c|ccc | ccc}\toprule
         & \multicolumn{3}{c|}{ARO~\cite{yuksekgonul2023when}} & \multicolumn{3}{c}{CREPE~\cite{ma2023crepe}} \\\hline
         Model & Attribute & Relation & Order & Atom & Swap & Negate  \\ \hline
        CLIP & 0.59 & 0.62 & \textbf{0.48} & 0.20 & \textbf{0.19} & \textbf{0.35}\\
        Ours & \textbf{0.65} & \textbf{0.65} & 0.45 &  \textbf{0.23} & \textbf{0.19} & 0.13\\\hline
        Gains & +10.1\% & +4.8\%& -6.3\% & + 15.0\% & 0.0\% & -31.5\% \\\toprule
    \end{tabular}}
    \caption{Comparison of our method with the baseline on ARO and CREPE datasets for text retrieval. Our method can discriminate texts which can be mapped to real scenes with different semantics (ARO-Attribute, ARO-Relation, CREPE-Atom) but struggles with linguistic phenomenons such as negation \textit{not} in CREPE, or grammatically incorrect sentences in ARO-Order.}
    \label{tab:aro}
\end{table}

\begin{figure}
    \centering
    \includegraphics[width=\linewidth]{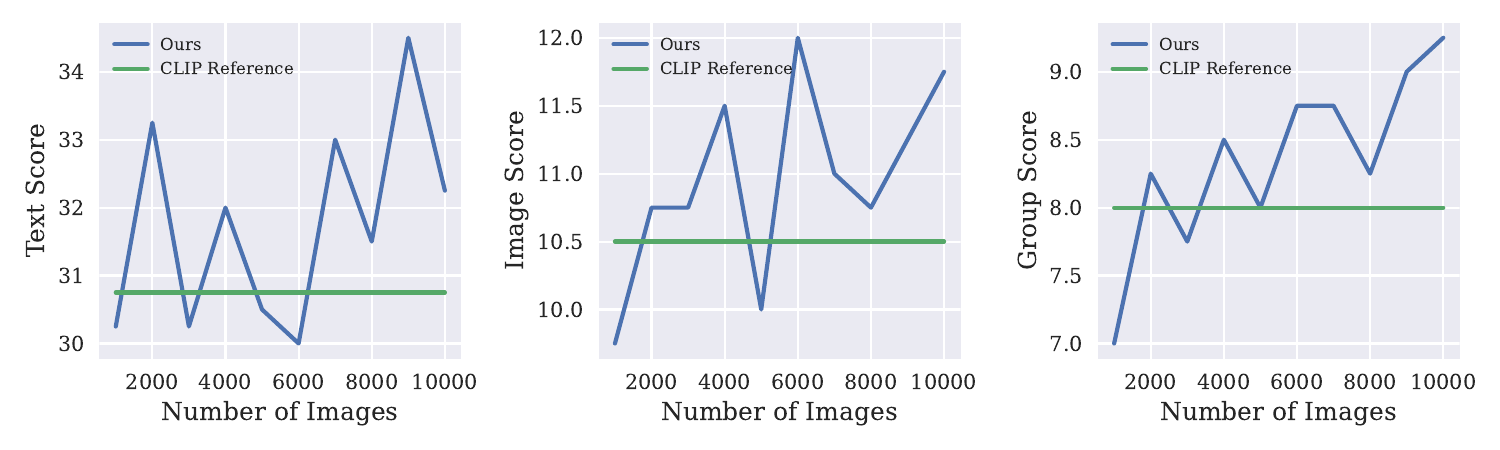}
    \caption{Model performance with increasing numbers of samples. Finetuning our model on incrementally increased generated data shows a consistent trend: as the data size grows, the model's performance is improved. This suggests the potential for generating more training examples to further enhance the model. Note the x-axis shows the number of unique images.}
    \label{fig:ablation-winoground}
\end{figure}

\begin{figure*}[htpb]
    \centering
    \includegraphics[width=0.95\linewidth]{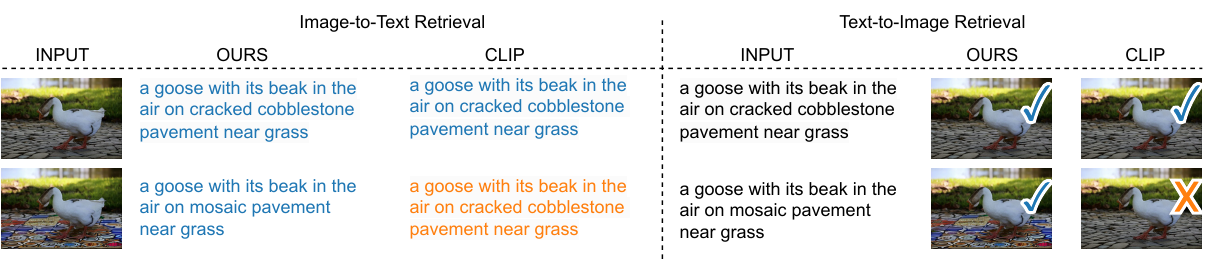}
    \caption{Examples of retrieved text for a given image (left) and retrieved images for a given text (right) by our method and the baseline. Correct matches are shown in blue, while the incorrect predictions are marked in orange.}
    \label{fig:qualitative}
    \vspace{-0.2in}
\end{figure*}

Table \ref{tab:aro} presents a comprehensive comparison of the performance between our CLIP and the baseline on the ARO and CREPE datasets. This analysis of the ARO dataset reveals an interesting phenomenon, wherein our method performs significantly better in the attribute category, e.g., matching the adjective \textit{white} to the object \textit{dog}, while demonstrating comparatively low performance in the order category, wherein the word order is randomly changed, e.g., \textit{a white cat} into \textit{cat a white}. This outcome is expected as our training data does not encompass sentences that are grammatically incorrect. 
Tab. \ref{tab:aro} confirms similar findings for CREPE. Our model demonstrates a significant improvement, especially in the atom category, where the objects and their attributes are changed. However, our model struggles with the negate category, such as transforming a dog into \textit{\textbf{not} a dog}. This outcome is expected as our training dataset lacks such examples.

\begin{table}[htpb]
    \centering
    \resizebox{0.95\linewidth}{!}{
    \begin{tabular}{c|ccc|c|c}\toprule
         & \multicolumn{3}{|c|}{VG} & SWIG & VAW \\
        Data & Object & Attribute & Relation & Object & Attribute \\ \hline
        CLIP & 79.0 & 69.8 & 58.2 & 71.8 & 65.7\\
        Ours & \textbf{85.1} &70.7 & 53.8 & 75.8  &  66.4 \\ \hline
        TSVLC~\cite{doveh2023teaching} & 82.8 & \textbf{75.5} & \textbf{62.6}  & \textbf{78.2} & \textbf{68.4} \\
       \toprule
    \end{tabular}}
    \caption{Comparison of our method with CLIP and TSVLC on VL-Checklist. The scores are obtained by averaging each subcategory within object, attribute, and relation.}
    \label{tab:vlchecklist}
\end{table}

Similar to previous findings, Table \ref{tab:vlchecklist} presents the improvements of our approach over CLIP on the VL-Checklist dataset in the object and attribute categories. The performance in the relation category is decreased as expected. Furthermore, we compare our method to a state-of-the-art approach proposed in TSVLC, which solely utilizes text augmentations. It is crucial to note that the comparison is not fair as the SOTA approach has been trained on a much larger dataset with a larger batch size and curated losses. Nevertheless, our model demonstrates comparable performance to that approach. A comprehensive analysis of detailed subsets of the VL-Checklist is in Appendix D.

\subsection{Ablation Study}

\textbf{The influence of human-curated dataset COCO.} Table \ref{tab:ablation-coco} provides a comparison between our generated dataset and the use of only the COCO dataset, which contains ground truth image text pairs from human annotators. The COCO image text pairs are part of our constructed dataset, and it was not clear if CLIP has incorporated COCO in its pretraining. To ensure the rigor of our analysis and examine the effect of existing labeled image-text pairs, we conduct an experiment comparing the performance of our method with CLIP that is finetuned only on the COCO dataset. Even though finetuning on COCO bring marginal improvement, our improvement is much more significant. This supports the assumption that the existing dataset may not contain sufficient hard negative examples.

\begin{table}[htp]
    \centering
    \resizebox{0.6\linewidth}{!}{
    \begin{tabular}{c|ccc}\toprule
        Model & Text & Image & Group  \\ \hline
        CLIP & 30.75 & 11.0 &  8.75 \\
        CLIP-COCO & 30.75 & \textbf{12.5} & 9.5\\
        Ours & \textbf{34.25} & \textbf{12.5} & \textbf{10.0} \\
       \toprule
    \end{tabular}}
    \caption{The influence of training data on the model performance on the Winoground dataset. CLIP-COCO is a finetuned CLIP model using our finetuning protocol on the COCO dataset. Ours denotes our final model finetuned on the mixture of our generated dataset and COCO.}
    \label{tab:ablation-coco}
\end{table}

\textbf{Impact of the number of generated samples.} In Fig. \ref{fig:ablation-winoground}, we analyze the impact of the data size of our generated samples on the Winoground performance. As seen from the figure, when the number of generated data samples increases, our model's performance on the image-text reasoning task improves. This demonstrates the value of incorporating more data in training to enhance the model's capabilities. This highlights the advantage of our method which utilizes large generative models, such as LLMs and text-to-image models, to generate high-quality examples, essential for tackling the vast image space. Due to hardware issues, our experiments are conducted until the data scale shown in the figure. Nonetheless, the results reveal a clear upward trend, indicating the potential for further improvements with a larger dataset.

\subsection{Qualitative Results}

Fig. \ref{fig:qualitative} shows a qualitative comparison of our model and the naive CLIP model on our test benchmark. For each input image or text, the most similar text or images among the two candidate texts or images are found by our method and CLIP. While our method can distinguish the details of the image, the CLIP model fails on this task. We report the performance of our models on our test set in Tab. \ref{tab:result-ours}.

\begin{table}[htp]
    \centering
    \resizebox{\linewidth}{!}{
    \begin{tabular}{c|ccc|ccc}\toprule
        Model & Text All & Image All & Group All&   Text 1 & Image 1 & Group 1  \\ \hline
        CLIP & 21.51 & 20.79 & 10.75 &60.21 &57.87 & 40.64\\
        Ours & \textbf{27.96} & \textbf{24.01} & \textbf{13.62} & \textbf{62.23} & \textbf{60.21} & \textbf{43.19} \\
       \toprule
    \end{tabular}}
    \caption{Comparison of our finetuned model to the CLIP baseline on our generated test set, evaluated in a similar metric as Winoground\cite{thrush2022winoground}.}
    \label{tab:result-ours}
    \vspace{-0.2in}
\end{table}

\section{Limitations}
Our work is limited by the performance of generative models such as ChatGPT and Stable Diffusion. We rely on the capacity of such models to produce high-quality examples. The diversity is mostly constrained by the power of LLMs. Additionally, we primarily manipulate local features such as objects and background, while may restrict the scope of negative examples generated, e.g., manipulating the relationship between two objects. Addressing these limitations is crucial for future improvements.

\section{Conclusion}
Our work tackles the limitations of existing visual language models in terms of compositional reasoning between text and images. We proposed a data generation pipeline that leveraged generative models to introduce challenging negative examples required for contrastive learning. Our proposed method effectively improves the compositionality and discriminative capabilities of VLMs. Experimental results demonstrate that training with our method consistently outperforms existing VLMs on various compositional reasoning benchmark datasets.  This was done by addressing the scarcity of hard negative examples for both the image and text modalities. Our work highlights the importance of generative approaches in advancing the field of visual language understanding and bridging the gap between humans and VLMs on compositional reasoning tasks.

\vspace{0.2cm}
\noindent\textbf{Acknowledgement}
This work was supported by the ERC Advanced Grant SIMULACRON.

\newpage

{\small
\bibliographystyle{ieee_fullname}
\bibliography{egbib}
}

\clearpage
\appendix

\section{Filtering Strategies}

\subsection{Failure Modes}

In our experiments, we conduct a manual evaluation of synthesized images and identify certain failure modes introduced below:
1. \textbf{object tag leads to wrong mask}: The prompted object is not correctly segmented from the context. This leads to an unusable generation or increased complexity as the original object remains unaffected.
2. \textbf{excessive segmentation}: In complex scenes, items are segmented towards part of the image that no longer contains the intended object. This degrades the image composition and may rendered the associated caption invalid.
3. \textbf{poor inpainting}: The performance of Stable Diffusion is affected by previous steps, including image complexity, portrayal quality.
4. \textbf{unusual state of the object}: In specific cases, objects appear in an unexpected positions and angles, making it challenging to inpaint the area.
5. \textbf{confusion due to multiple instances}: When multiple items are present in an image, the generation performance may be decreased. For example, in an image of multiple plates, painting one of them does not effectively change the meaning as expected.
6. \textbf{high complexity in the image}: Images with a variety of objects may cause small portions of the image left for each of them.
7. \textbf{small mask size}: In some cases, the identified object is so small that generation fails due to poor quality.
8. \textbf{lack of descriptiveness in portrayal}: ChatGPT may produces portrayals unsuitable for image generation. The lack of descriptiveness can lead to nearly identical images with minimal differences.
9. \textbf{animate objects}: Animals and humans are hard to portray because their posture dynamically changes with the action.

\begin{figure}[htp]
    \centering
    \includegraphics[width=\linewidth]{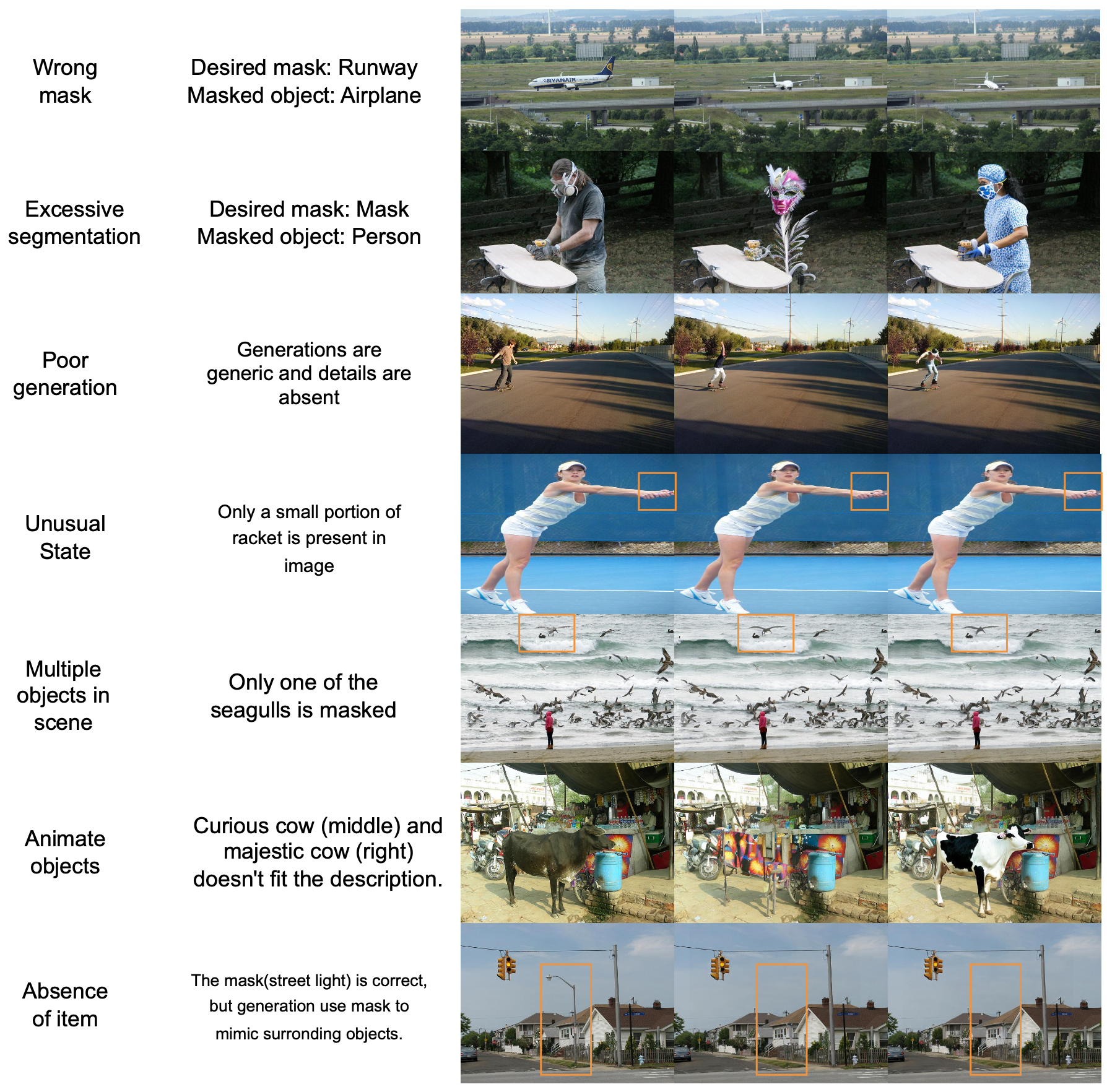}
    \caption{Samples of detected failure modes. The left side shows original images whereas the middle and right sides show two generations.}
    \label{fig:fig3}
\end{figure}

Figure \ref{fig:fig3} illustrates the failure modes in image generation, where original and generated versions of images are presented.

\subsection{Implementation of Filters}

To address these issues, we mainly use two filters. The first filter is the BLIP ITM head that returns a matching score between an image and text. The second filter calculates the variance within object-level generations.

\textbf{ITM Filter}
BLIP’s ITM head outputs a confidence value if a given image and text pair match. As suggested in their original work, we use the decision threshold of 0 to pass a sample as valid. We utilize BLIP to compute two scores, the variation score that grades the inpainted image and patched caption, and the original score that grades the inpainted image and original caption. The variation score is used in our filtering pipeline. The original score is only used for statistical analysis, as it is not capable of determining fine-grained concept matching.

\begin{figure}[htp]
    \centering
    \includegraphics[width=\linewidth]{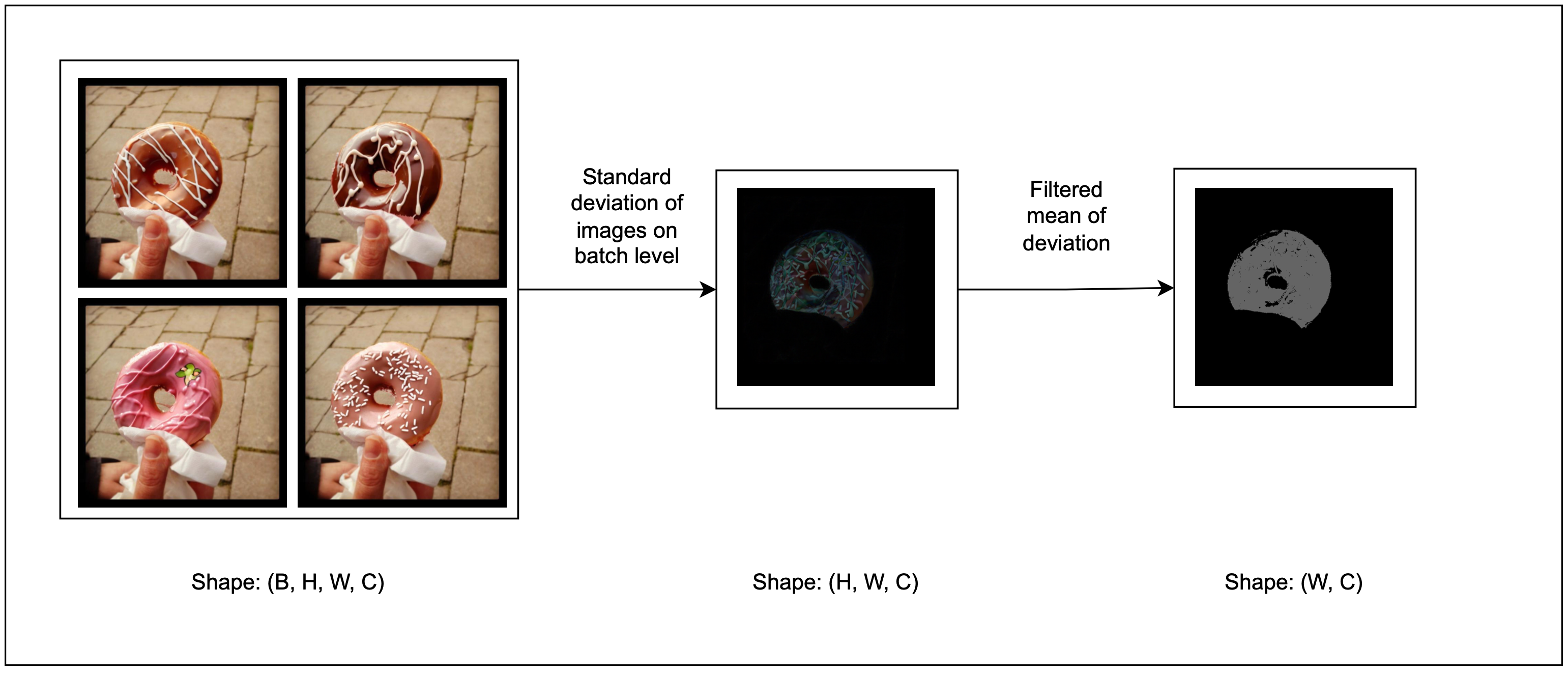}
    \caption{Calculation of Image Variation Area, visualized}
    \label{fig:fig4}
\end{figure}

\textbf{Variance Filter}
To filter out samples with small mask sizes and samples generated similarly, we calculate a variance score for images, as shown in Figure \ref{fig:fig4}. First, we stack all the images and create a tensor of size (Batch Size, Channels, Height, Width). Then we calculate the standard deviation of the first dimension, in which we know the area outside of the mask should yield 0 since they have identical values. For the masked part, we further calculate the mean of the channels, reducing the tensor to shape (Height, Width). In this form, we apply a condition whether the average deviation is higher than a value $\epsilon$. The final array is a boolean array, in which we average across two dimensions to calculate the latest value. The next section provides a deeper insight into the statistics of the filtering process.

\subsection{Analysis of Generated Images}

\begin{figure}[htp]
    \centering
    \includegraphics[width=\linewidth]{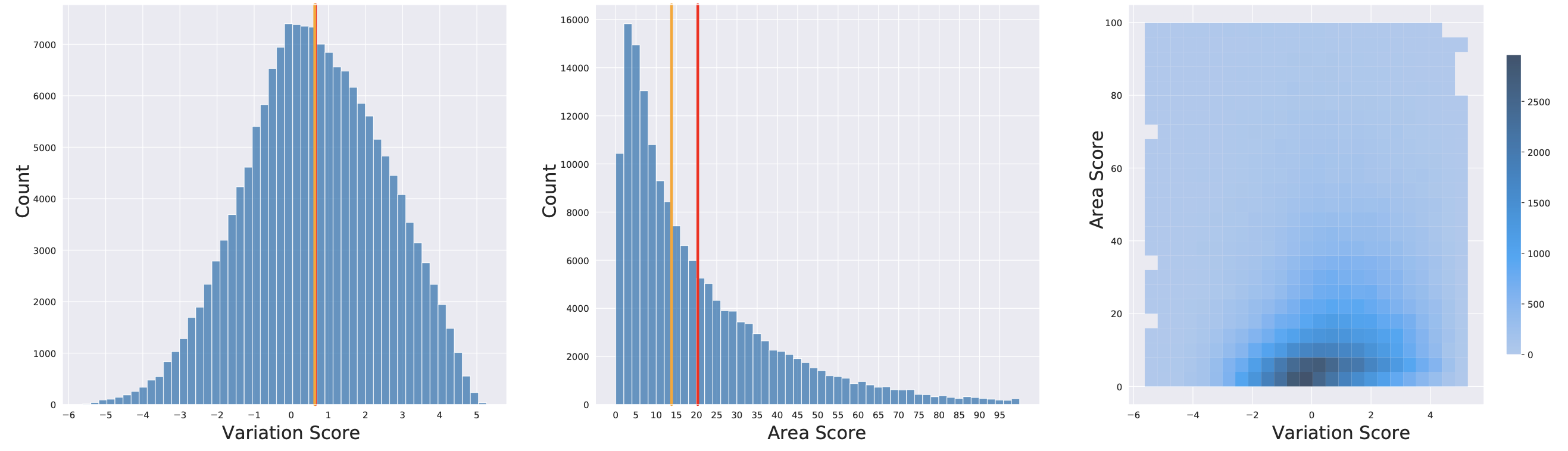}
    \caption{2D histogram of Variation Score and Area Score. The red line indicates the mean and the orange line indicates the median. We observe normal distribution on ITM Score and log-normal distribution on the area score. We prefer ITM Score $>$ 0 and Area Score $>$ 14}
    \label{fig:fig5}
\end{figure}

To gain deeper insights into generated variation images, we analyze the histogram of filter scores in Figure \ref{fig:fig5}. As shown in the left histogram of the figure, the variation score ranges between -6 and 5. Our manual examination reveals that the samples with scores between -1 and 0 contain optimal images. Although this range may result in a loss of high-quality generations, we did not dive deeper into mining such samples. However, advanced filtering can be utilized to improve the generation. The middle histogram reveals that a significant amount of samples exhibit a small image variation area. Therefore, we use the median value and set the threshold for filtering to 14.

Another metric related to the variation area is the delta in the mask. Within each variation group that replaces the same object, we calculate a delta score inside the boundaries of the mask. The delta in mask calculation is similar to image variation calculation. However, we skip the step to check if the value is larger than the epsilon value to measure the amount of difference. We employ this value for our statistical analysis. Items with high delta and high masking percentages tend to have generations aligned with their portrayals. Objects with greater size in physical life are likely to have higher masking percentages. However, there are also items like pizza which is relatively small in physical life compared to other objects but still covers a higher area in images.

\begin{figure}[htp]
    \centering
    \includegraphics[width=\linewidth]{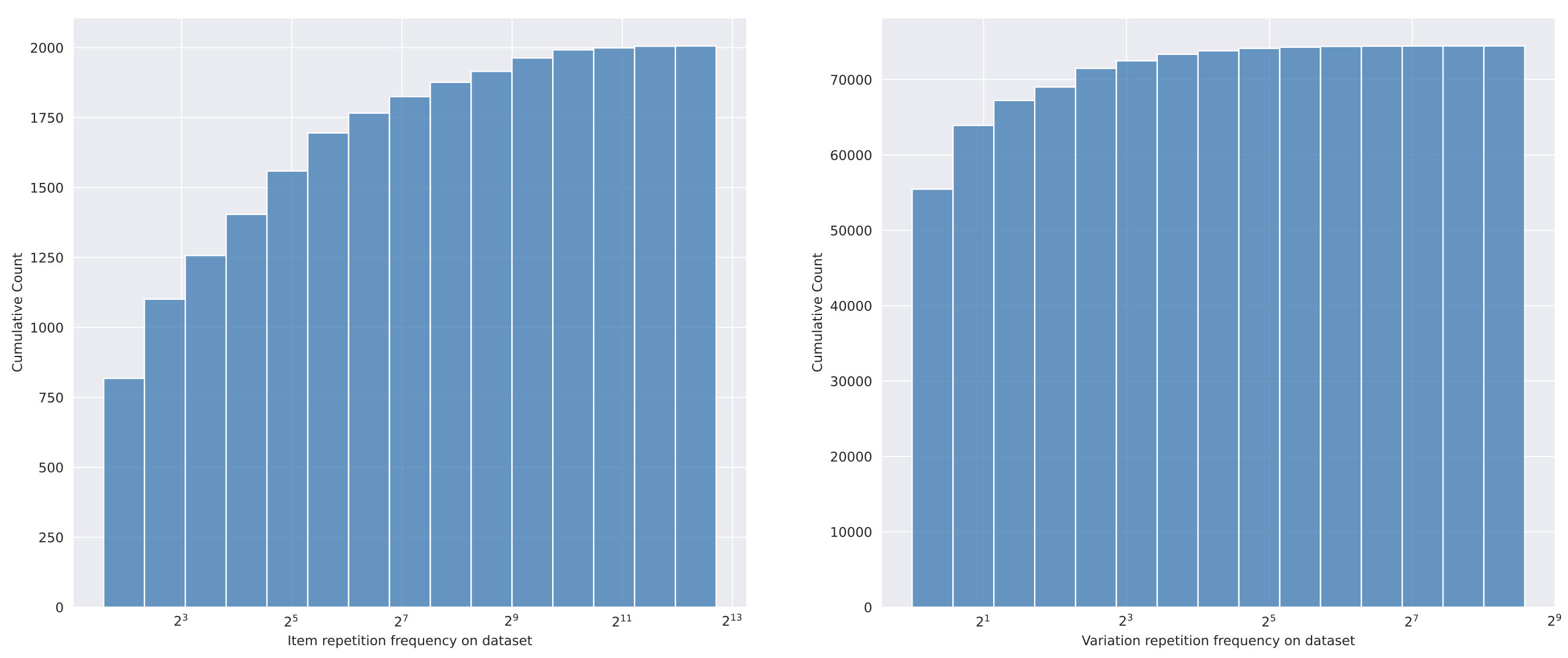}
    \caption{Repetition frequency of items and variations}
    \label{fig:fig7}
\end{figure}

\begin{figure*}[htp]
    \centering
    \includegraphics[width=0.7\linewidth]{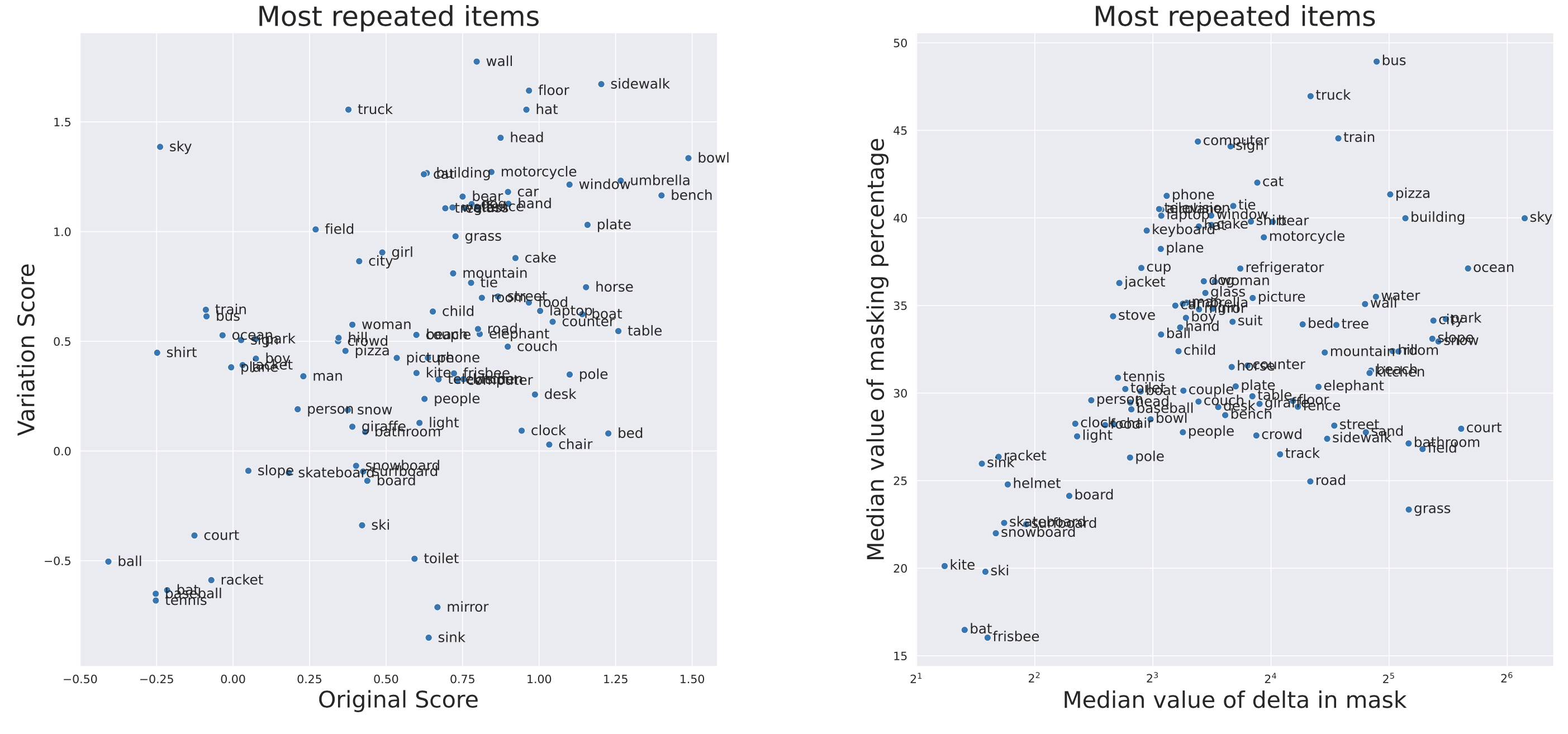}
    \caption{Most common items by their average scores of filters. In the left plot, the x-axis represents the median value of the original score. The y axis represents the median value of variation scores. In the right plot, the x-axis represents the median value of the delta in the mask while y-axis represents the median value of the masking percentage.}
    \label{fig:fig6}
\end{figure*}

An analysis of filters on the object level provides further insight into the statistics of the objects. We visualized the most commonly portrayed items after filtering in Figure \ref{fig:fig6}.

The training dataset includes 164,021 variations generated for 41,003 objects identified in 12,656 images. There are 2006 different items and 74,460 different phrases. ChatGPT utilizes a total number of 9,464 unique tokens (identified by the CLIP tokenizer) and 12,413 unique words. Objects like heads and pictures have the most diverse set of portrayals. Common objects categories such as keyboards, desks, trains, and kitchens have the most repetitive representations. We observe a uniqueness drop with respect to the total count. The examination of highly repetitive objects that the words used in portrayals are not descriptive enough. For example, "minimalist kitchen" and "modern kitchen" are two frequent responses ChatGPT has produced for the kitchen. Nevertheless, minimalist or modern does not express a niche vision but a general concept.

Last, we check item frequency and uniqueness of short descriptions. As presented in the left histogram of Figure \ref{fig:fig7}, the count of items identified only once is 818. A glance over singular items reveals their infrequent existence. The majority of singular objects are uncommon in everyday environments: asparagus, octopus, orchid, etc. A fractional subset of the singular objects are phrases: a herd of giraffes, mountain goat, car mirror. This indicates that the COCO dataset doesn’t have a diverse set of objects present in the pictures. The right side of Figure \ref{fig:fig7} presents the frequency histogram of short descriptions. The count of variations identified only once is 55,458. However, there are more than 200 occurrences of short descriptions like a cobblestone street, a sleek glass table, and a majestic oak tree. Such high repetitions indicate ChatGPT is repetitive in terms of variation generations.

\section{Evaluation on Winoground Splits}
As mentioned in Section \textcolor{red}{5.1}, we evaluate our model's performance on the splits of Winoground that test different aspects of reasoning capabilities. From Tab. \ref{tab:supp}, we see that our model outperforms the baseline in the compositional reasoning task which requires detailed descriptions of visual scenes. However, our model fails when comprehending certain difficult texts, especially when the meaning of the text becomes challenging. For example, in the case of the phrase \textit{the brave in the face of fear}, the image depicts a small cub confronting a fierce lion, while the model needs an in-depth understanding of the word \textit{brave} to associate it to the cub. During the finetuning process, our model may demonstrate the phenomenon of "catastrophic forgetting", if the quality, diversity, and scale of our dataset do not match with the original pretraining dataset. In particular, the presence of repetitive text samples in our augmented dataset may impede the performance of the text encoder.

\begin{table}[htp]
    \centering
    \scriptsize
    \begin{tabular}{c|ccc|ccc}\toprule
        & \multicolumn{3}{c|}{Compositional (171)} & \multicolumn{3}{c}{Complex (78)} \\\hline
       CLIP & 31.58&            11.70&           9.36& 23.08&            6.41&           3.85\\
        Ours & \textbf{38.01}&           \textbf{14.62}&           \textbf{10.53} & \textbf{29.49}&  \textbf{8.97}&   \textbf{6.41}\\ \hline
        Gains &  +22.5\% & +27.2\% & +12.5\% & +23.9\% & +39.9\% & +66.5\% \\ \toprule
        & \multicolumn{3}{c|}{Unusual Image (56)} & \multicolumn{3}{c}{Unusual Text (50)} \\\hline
        CLIP & 26.79&           \textbf{8.93}& 5.36& \textbf{34.0}&            \textbf{14.0}&            \textbf{10.0}\\
        Ours & \textbf{28.57}&           \textbf{8.93}&           \textbf{8.93}& 30.0&            10.0&            \textbf{10.0}\\ \hline
        Gains & +6.7\% & 0.0\% & +66.3\% & -11.8\% & -28.5\% & 0.0\% \\ \toprule
        & \multicolumn{3}{c}{Ambiguous(46)} & \multicolumn{3}{c}{Visually Difficult(38) } \\ \hline
        CLIP & \textbf{30.43}&           \textbf{15.22}&           \textbf{15.22}& 15.79&   0.00 & 0.00\\
        Ours & 26.09&           8.70&           8.70& \textbf{18.42}&  \textbf{2.63}&  \textbf{2.63}\\ \hline
        Gains & -14.2\% & -43.8\% & -43.8\% & +16.6\% & +2.63\% & +2.63\% \\ \toprule
         & \multicolumn{3}{c|}{Non compositional(30) } \\ \hline
         CLIP & \textbf{76.67}&           36.67&           33.33&\\
         Ours & 70.00&            \textbf{40.00}&           \textbf{36.67}&\\
         Gains & -8.7\% & +9.0\% & +10.0\%\\
        \toprule
    \end{tabular}
    \caption{Comparison of models on Winoground subsets that evaluate distinct reasoning abilities. The numbers in parentheses represent the sample count for each split. Our model excels in compositional reasoning tasks that demand a detailed description of the scene. However, it struggles when it comes to understanding subtle differences in the text that may require background knowledge, e.g., unusual text.}
    \label{tab:supp}
\end{table}

\section{Examples from Our Dataset}
\begin{figure*}
    \centering
    \includegraphics[width=\linewidth]{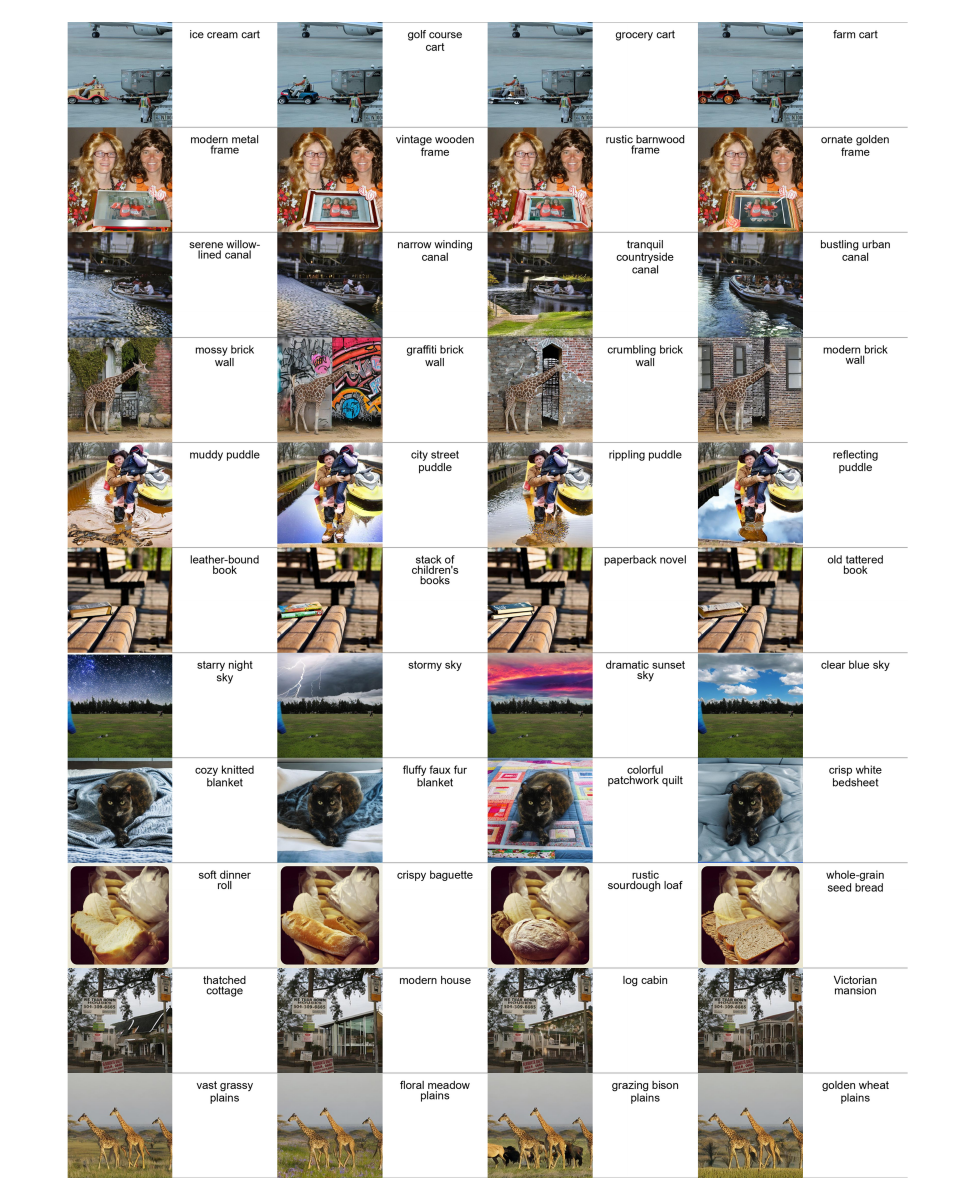}
    \caption{Examples from our generated dataset. Each row demonstrates four variations generated using a COCO image-text pair. To highlight the differences among the images, we only provide descriptions for the modified part, instead of the caption for the entire image.}
    \label{fig:supp}
\end{figure*}

In Fig. \ref{fig:supp}, we showcase a few examples from our generated dataset. Our approach is advantageous in that we can generate a diverse dataset with challenging negative examples. For instance, the images in the first row depict scenarios that are highly unlikely in the real world, since an ice cream cart will never appear at an airport for aircraft maintenance. These examples serve as a true test of the model's understanding of the cart concept.

Furthermore, some examples in our dataset differ in fine-grained details that can be challenging even for humans. An example of this can be observed in the last row. The model needs to analyze the specific type of grass in order to make an accurate prediction.

\section{Comparison with SOTA Method}
To obtain a comprehensive understanding of the compositional reasoning ability of our approach, we conduct a comparative analysis with a state-of-the-art method, TSVLC~\cite{doveh2023teaching}, on two established benchmarks, Winoground and VL-Checklist. Table \ref{tab:supp-wino} presents the evaluation results on Winoground, where our model significantly outperforms TSVLC in terms of text score, despite lower image score and group score. For VL-Checklist, we present the evaluation results on detailed data subsets\footnote{We were unable to download the full HAKE dataset due to a server failure. Nevertheless, the evaluation of existing datasets provides representative insights into the compositional reasoning capabilities of our model.} in Table \ref{tab:supp-vl-vgo}, \ref{tab:supp-vl-vgr}, \ref{tab:supp-vl-swig}, and \ref{tab:supp-vl-vaw}. We obtained the evaluation scores of TSVLC from their published paper. From the results presented in the table, we find that our model performs comparably to TSCVL (The average of all individual metrics in Table \ref{tab:supp-vl-vgo}-\ref{tab:supp-vl-vaw} yields the following results: CLIP: 70.57, Ours: 72.37, TSVLC: 75.71). Note that TSCVL is trained on 3 million image-text pairs, while our approach is finetuned on a much smaller scale of approximately 100k images. In addition, TSCVL incorporates more sophisticated negative sampling strategies and curated loss functions. In contrast, we utilize the naive CLIP architecture and the simple contrastive loss function. Furthermore, it is worth noting that our method still outperforms CLIP on average, consistent with the observation on other datasets presented in the main text.

\begin{table}[htp]
    \centering
    \resizebox{0.85\linewidth}{!}{
    \begin{tabular}{c|ccc}\toprule
        Model & Text Score & Image Score & Group Score  \\ \hline
        CLIP & 30.75 & 11.0 &  8.75 \\
        TSVLC & 26.0 & \textbf{15.75 }& \textbf{11.0}\\
        Ours & \textbf{34.25}&  12.5& 10.0  \\ 
       \toprule
    \end{tabular}}
    \caption{Comparison of our method with CLIP and TSVLC on Winoground benchmarks. We report the text score, image score, and group score which measure if the model can correctly match a text for an input image, or vice versa.}
    \label{tab:supp-wino}
\end{table}

\begin{table*}[htp]
    \centering
    \begin{tabular}{c|ccccccc}\toprule
    &O-Large &O-Medium& O-Small& O-Center &O-Mid& O-Margin &Avg O \\\hline
CLIP & 86.95 & 77.75 & 72.75 & 85.5 & 80.5 & 70.6 & 79.00 \\
TSVLC & 90.5 & 81.95 & \textbf{77.6} & 89.75 & 83.8& 73.35 & 82.82\\
Ours & \textbf{92.04} & \textbf{84.97} & 77.52  & \textbf{92.01}  & \textbf{85.95} & \textbf{77.99} & \textbf{85.08} \\ \toprule
    \end{tabular}
    \caption{Evaluation on VG Object subset of VL-Checklist. TSVLC refers to the finetuned model on CC3M. Our approach utilizes CLIP and is further finetuned on our augmented dataset. Our model outperforms both the CLIP and TSVLC approaches.}
    \label{tab:supp-vl-vgo}
\end{table*}

\begin{table*}[htp]
    \centering
    \begin{tabular}{c|cccccccc}\toprule
         &A-Color & A-Material & A-Size & A-State & A-Action & R-action & R-spatial & Avg A+R\\\hline
        CLIP & 68.9 & 65.4 & 72.1 & \textbf{69.3}& 72.37 & \textbf{62.4} & 54.0 & 66.35 \\
        TSVLC & \textbf{79.9} & \textbf{78} & \textbf{76.8} & 68.7 & 74.18 & 61.9 & \textbf{63.2} & \textbf{71.81} \\
        OUrs & 73.07 & 72.51 & 64.38 & 67.91 & 75.53 &  57.86 & 49.69 & 65.85 \\ \toprule
    \end{tabular}
    \caption{Evaluation on VG Attribute and Relationship subset of VL-Checklist. TSVLC refers to the finetuned model on CC3M. Our approach utilizes CLIP and is further finetuned on our augmented dataset.}
    \label{tab:supp-vl-vgr}
\end{table*}

\begin{table*}[htp]
    \centering
    \begin{tabular}{c|ccccccc}\toprule
              & O-Large & O-Medium & O-Small & O-Center & O-Mid & O-Margin & Avg All\\ \hline
        CLIP & 76.98 & 73.28 & 59.41 & 78.075 & 74.63 & 64.49 & 71.76 \\
        TSVLC & \textbf{83.5} & \textbf{80.05} & \textbf{71.70} & \textbf{84.02} & \textbf{81.17} & \textbf{75.01} & \textbf{78.24} \\
        Ours & 81.11 & 75.04 & 68.82 & 81.45 & 78.00 & 70.53 & 75.82\\\toprule
    \end{tabular}
    \caption{Evaluation on SWIG subset of VL-Checklist. TSVLC refers to the finetuned model on CC3M. Our approach utilizes CLIP and is further finetuned on our augmented dataset. Though our model underperforms TSVLC, it still exhibits significant improvement over CLIP.}
    \label{tab:supp-vl-swig}
\end{table*}

\begin{table*}[htp]
    \centering
    \begin{tabular}{c|cccccc}\toprule
       & A-Color & A-Material &A-Size &A-State &A-Action &Avg All \\ \hline
        CLIP & 71 & 73.3 & 68 & 53.3 & 62.7 & 65.66\\
        TSVLC & \textbf{75} & \textbf{76.7} & \textbf{69.9} & 55.9 & 64.6 & \textbf{68.42} \\
        Ours& 76.6 &  68.7 & 56.23 &  57.99 &  72.62 & 66.4 \\\toprule
    \end{tabular}
    \caption{Evaluation on VAW subset of VL-Checklist. TSVLC is the final model which is finetuned on CC3M. Our approach utilizes CLIP and is further finetuned on our augmented dataset.}
    \label{tab:supp-vl-vaw}
\end{table*}

\end{document}